\documentclass[journal]{IEEEtran}
\usepackage{amsmath,bm}

\usepackage{cite}
\usepackage{multirow}
\usepackage{diagbox}
\usepackage{lipsum}
\usepackage{graphicx}

\usepackage{enumitem}
\usepackage{amssymb}
\usepackage{booktabs}

\usepackage{hyperref}
\usepackage{color,soul}
\usepackage[dvipsnames]{xcolor}

\usepackage[ruled,linesnumbered]{algorithm2e}

\ifCLASSOPTIONcompsoc
\usepackage[caption=false,font=normalsize,labelfon
t=sf,textfont=sf]{subfig}
\else
\usepackage[caption=false,font=footnotesize]{subfi
g}
\fi
\usepackage{alphalph}
\usepackage{tabularx,ragged2e}
\usepackage{threeparttable}
\usepackage{hyperref} %<--- Load after everything else

\ifCLASSINFOpdf
\else
\fi
\begin{document}
\title{Language bias in Visual Question Answering: A Survey and Taxonomy}
% \author{Michael~Shell,~\IEEEmembership{Member,~IEEE,}
%         John~Doe,~\IEEEmembership{Fellow,~OSA,}
%         and~Jane~Doe,~\IEEEmembership{Life~Fellow,~IEEE}% <-this % stops a space
% \author{Zhengzhong~Tu\textsuperscript{\orcidicon{0000-0002-7594-2292}}, 
%         Yilin~Wang,
%         Neil~Birkbeck,
%         Balu~Adsumilli,
%         and~Alan~C.~Bovik,~\IEEEmembership{Fellow,~IEEE}% <-this % stops a space

\iffalse
\author{Desen~Yuan,~Qingbo~Wu,~King~N.~Ngan,~Hongliang~Li,~Fanman~Meng,~and~Linfeng~Xu

}

\thanks{Q. Wu, D. Yuan, K. N. Ngan, H. Li, F. Meng and L. Xu were with the School
of Information and Communication Engineering, University of Electronic Science and Technology of China, Chengdu,
611731 China e-mail: (qbwu@uestc.edu.cn;desenyuan@gmail.com;knngan@uestc.edu.cn;hlli@ue-stc.edu.cn;fmmeng@uestc.edu.cn;lfxu@uestc.edu.cn).}

\fi

\author{Desen~Yuan

\thanks{D. Yuan were with the School
of Information and Communication Engineering, University of Electronic Science and Technology of China, Chengdu. This is the first version of this survey. There are some grammatical and detailed problems. Due to the author's time, the survey will be updated in recent months.}}

\maketitle

\begin{abstract}
Visual question answering (VQA) is a challenging task, which has attracted more and more attention in the field of computer vision and natural language processing. However, the current visual question answering has the problem of language bias, which reduces the robustness of the model and has an adverse impact on the practical application of visual question answering. In this paper, we conduct a comprehensive review and analysis of this field for the first time, and classify the existing methods according to three categories, including enhancing visual information, weakening language priors, data enhancement and training strategies. At the same time, the relevant representative methods are introduced, summarized and analyzed in turn. The causes of language bias are revealed and classified. Secondly, this paper introduces the data sets mainly used for testing, and reports the experimental results of various existing methods. Finally, we discuss the possible future research directions in this field.
\end{abstract}


% \fi
\IEEEpeerreviewmaketitle

\section{Introduction}

\IEEEPARstart{V}{isual} question answering~\cite{antol2015vqa,agrawal2017vqa} is a cross task combining computer vision and natural language processing. It is a hot task in the application and research of deep learning. Computer vision task studies how to let the machine process and understand the content in the image. Its main research contents include object detection\cite{ren2016faster, he2017mask, lin2017feature, lin2017focal,ding2020human,qiu2020hierarchical,qiu2020offset,wang2020multi,li2020codan,li2019simultaneously,li2019headnet,qiu2019a2rmnet,chen2021bal,chen2020high,li2018incremental }, action recognition~\cite{Carreira2017Quo,tran2018closer, tran2015learning}, image classification\cite{deng2009imagenet, he2016deep, he2017mask}, image segmentation\cite{he2017mask,yang2020new,guo2020deep,shi2020query,yang2020learning,yang2020mono,meng2019new,yang2019new,xu2019bounding,shang2019instance,luo2018weakly,shi2018key}, image dehazing\cite{wei2020single,li2020region,wu2020subjective,li2020haze,wu2020unified,luo2021single,wu2019beyond,wei2021non,li2021single} and so on \cite{song2017letrist,ma2016waterloo,song2013noise,ma2016group,wu2017blind,meng2013feature,wu2015no,wu2017blind1}. Natural language processing (NLP) is a process that allows machines to analyze and understand human language, including articles, sentences and emotions. Its main research tasks include machine translation\cite{bahdanau2014neural, papineni2002bleu}, named entity recognition\cite{curran2003language, lample2016neural}, emotion recognition\cite{huang2019character, cowie2001emotion, ghosal2019dialoguegcn} and so on. Visual question answering belongs to the cross field of these two research fields, namely multimodal machine learning. Visual question answering is a downstream task of multimodal machine learning.

In the application and research of multimodal machine learning, visual question answering is becoming more and more important. In the past decades, significant research progress has been made in computer vision and natural language processing. The explosive growth of visual and text data that can be obtained and processed makes these two fields develop rapidly. At present, the widely used and proven effective methods in computer vision and natural language processing include convolution neural network, cyclic neural network, long-term and short-term memory network and transfmer. These methods have good applications in visual problem tasks.

In the most common form of visual question and answer (VQA), data includes a picture and a question, which requires the machine to give the corresponding answer to its question. The main difference from other computer vision tasks is that the questions to be answered by the model change in real time, while in other computer vision tasks, such as target detection and target segmentation, they are given in advance. Only the image changes, while the image and text of VQA task change in real time. VQA task is more in line with the real form of artificial intelligence, which can help the model to understand vision and language more deeply. The answer can be searched in a given text set or external knowledge base. Compared with the pure text answer (QA) task, VQA task adds image data containing more information. However, due to the noisy information of image data and the lack of information structure and syntax rules that can be simply described, it is very different from text data.

Therefore, VQA can be regarded as a real and perfect application of artificial intelligence technology, which requires multi-domain and multi-modal knowledge. However, due to the bottleneck of task evaluation, the current VQA task still has great limitations. The generated answer is usually relatively simple, mostly a few words or simple sentences, rather than longer sentences. VQA task is still a challenging and open research problem.

In the field of visual question answering, a hot research problem in recent years is how to solve the problem of language bias. For example, for many visual question answering models, the answer can be inferred directly only through language data. Bias in VQA generally refers to   Language bias, which makes the model rely on the surface correlation between the question and the answer when answering the question, and ignores the image information. A classic example is that for the question "what color is the banana in the figure?", although the banana in the figure is immature "green", the model still tends to predict "yellow".

Language bias causes great difficulties in the application of VQA. Language bias can be decomposed into language shortcut bias generated by the model or language data bias caused by problem data distribution. Due to the existence of language bias, the model will have poor robustness in application, and the interpretability of the model is very low. Similar to language bias, this bias often exists widely in multi-modal data, which seriously affects the practical application of multimodal machine learning. At the same time, it may lead to discrimination. For example, deploying relevant VQA models in public places such as social platforms may produce some columns of discrimination and misleading due to the bias, affecting the fair use of technology.

In order to solve the problem of language bias, a large number of papers on how to solve language bias have appeared in the past few years. The purpose of this survey is to give a comprehensive overview of this field, including models, data sets, etc., and put forward the possible future direction of this field. As far as we know, this paper is the first survey on language bias in VQA.

In the first part of this survey (Chapter 2), according to the current methods to solve language bias, this paper divides the existing methods into three categories: enhancing visual information, weakening language a priori, data enhancement and training strategies. At the same time, the basic network architecture of existing methods is also introduced. In the third chapter, this paper introduces the relevant information of the current mainstream data sets and the relevant evaluation schemes, and provides the experimental results of the existing methods. In Chapter 4, we synthesize the current methods and the core problems of language bias, and give the possible development direction of this field in the future. In Chapter 5, we review the content of this paper.

\section{Methods for debiased VQA}
\label{sec:ugc_db}

\subsection{Backbones}

In order to solve the multi class classification problem in the field of VQA, researchers have proposedd many basic models to be applied to VQA. The general form of VQA is: Given a dataset $\mathcal{D}=\left\{I_{i},Q_{i},a_{i}\right\}^N$ containing N  triplets of images $I_{i} \in \mathcal{I}$ , questions $Q_{i} \in \mathcal{Q}$ and answers $a_{i} \in \mathcal{A}$.

The aim of the VQA task is to learn a mapping function $f_{vqa}$:$I \times Q \to [0,1]^{\left| \mathcal{A}\right|}$, which generates the answer distributions for any given image-question pairs.

\noindent\textbf{\emph{GVQA:}} GVQA divides the task of VQA into two steps, Look: find the object / image block required to answer the question and identify the visual concept in the block; Find the space for reasonable answers from the questions, and return the correct visual concepts from a group of visual concepts by considering which concepts are reasonable. The novelty of GVQA is that answering "yes" or "no"is an independent task. GVQA includes 1) visual concept classifier (VCC): it is responsible for locating the image blocks required to answer questions and generating a set of visual concepts related to the location blocks. 2) Answer clustering predictor (ACP) 3) concept extractor (CE) 4) the outputs of VCC and ACP are fed to answer predictor (AP) to generate answers.

\noindent\textbf{\emph{SAN:}} Yang et al. proposed the SAN model. The model takes the problem as a query condition and looks for the area related to the problem in the figure. The whole model is divided into three modules: 1) image model: using VGGNet to extract image features, the selected features are the features of the last layer - the pool layer, which well maintains the spatial information of the original image. 2) Question model: extract text features using LSTM or CNN. 3) Stacked attention networks: realize the attention of image area through multiple iterations. Firstly, a feature attention distribution is generated according to image features and text features, and the weight of each region of the image is obtained according to this distribution. This process is iterated for many times, and finally the problem related region is noticed.

\noindent\textbf{\emph{UpDn:}} Anderson et al. Proposed a top-down and bottom-up attention model method, which was applied to the related problems of visual scene understanding and visual question answering system. The bottom-up attention model (using fast r-cnn) is used to extract the region of interest in the image and obtain the object features; The top-down attention model is used to learn the weight corresponding to the feature (using LSTM), so as to realize the in-depth understanding of the visual image. UpDn model won the first place in VQA 2017 challenge and is widely used to solve the problem of language bias.

\noindent\textbf{\emph{NSM:}} Manning et al. have introduced the Neural State Machine, a graph-based network that simulates the operation of an automaton, and demonstrated its versatility, robustness and high generalization skills on the tasks of real-world visual reasoning and compositional question answering. By incorporating the concept of a state machine into neural networks, they introduce a strong structural prior that enhances compositinality both in terms of the representation, by having a structured graph to serve as the world model, as well as in terms of the computation, by performing sequential reasoning over such graphs.

\noindent\textbf{\emph{MUREL and S-MRL:}} Cadène et al. introduced MuRel, a multimodal relational network for Visual Question Answering task. This system was based on rich representations of visual image regions that are progressively merged with the question representation. They included region relations with pairwise combinations in the fusion, and the whole system can be leveraged to define visualization schemes helping to interpret the decision process of MuRel. They validated the approach on three challenging datasets: VQA 2.0, VQA-CP v2 and TDIUC. They clearly demonstrating the gain of the vectorial representation to model the attention. S-MRL stands for Simplified-MUREL. The architecture was proposed in RUBi by Cadene et al.

\noindent\textbf{\emph{CAM:}} Peng et al. proposed a novel VQA model, termed Cascaded-Answering Model (CAM), to take advantage of the semantics of the predicted answers that are neglected by previous models. CAM extended the conventional one-stage VQA model to a two-stage model using two cascaded answering modules, Candidate Answer Generation (CAG) module and Final Answer Prediction (FAP) module.,CAG generated the candidate answers, and FAP predicted the final answer by combining the information of question, image and candidate answers.

\noindent\textbf{\emph{LXMERT:}} Tan et al. proposed a cross modal framework LXMERT, which uses a two-way transformer structure and is constructed based on transformer encoder and a new cross modal encoder. Three encoders are used to endow the model with cross modal capability through five pre-training methods. The encoders are object relationship encoder, language encoder and cross modal encoder. The model has two kinds of input information, one is the image-based object feature information, and the other is the text-based question embedding information.

\subsection{Strengthening visual information}
\label{ssec:observation}

Selvaraju et al. proposed the human importance perception network adjustment (HINT) , a general framework for aligning network sensitivity with spatial input regions that humans perceive as relevant to a task. The authors demonstrate the effectiveness of this approach in improving the visual basis of visual and language tasks such as VQA and image captioning. The research shows that the good grounding improves the generalization ability of the model and the credibility of the model.

Li et al proposed a new visual question-and-answer framework called relational graph attention network (REGAT) to model multi-type object relationships using question adaptive attention mechanism. Regat utilizes two types of visual object relations: explicit and implicit, in order to perceive region representation through graph-based attention learning, the new results are obtained on both VQA 2.0 and VQA-cp V2 datasets. Comprehensive experiments on two VQA data sets show that the model can be injected into the most advanced VQA architecture in a plug-and-play way.

Shrestha et al. advocate the use of synthetic data sets to create a basic fact-based data set for 100 percent of instances, allowing communities to assess whether their methods can focus on relevant information, another option is to use tasks that have a clear testing basis, for example, in visual query detection, the agent must be around the output box in any area of the scenario that matches the natural language query (Acharya et al. , 2019a) . Here, Shrestha et al. show that existing vision-based VQA bias mitigation methods do not work as expected.

Teney et al. suggested taking a random subset of 8K instances from VQA-CP training data to evaluate domain performance and compare it to some existing work. VGQE was proposed by KV et al. . It is a new type of problem encoder, which uses pattern and generates visual problem representation. This problem indicates sufficient discrimination based on visual counterparts and helps the model to reduce language bias from training sets. The VQA-CP V2 data set, which is sensitive to deviation, has been extensively tested by KV et al, and a new and up-to-date technology has been realized. Unlike existing deviation reduction techniques, VGQE does not sacrifice the performance of the original model on the standard VQA V2 benchmark.

Hirota et al. explored the use of text representations of images, rather than the deep visual features of VQA. In order to increase the size and diversity of training data, the authors explore the data expansion methods of description and problem. Through experiments, including ablation experiments, the author verifies the competitiveness of the pure language model and the model based on depth vision features. Most of the data enhancement techniques improve performance; in particular, problem-based back translation is a strong driver.

Si et al proposed a progressive framework of selection and reordering (Sar) based on visual implication, in which the authors first select candidate answers to reduce the prediction space, and then reorder the candidate answers through visual inclusion tasks, the task verifies that the image semantically contains a composite statement of the question and each candidate answer. The framework can take full advantage of the interaction of images, questions, and candidate answers. It is a general framework that can be combined with existing VQA models to further enhance their capabilities. Si et al. demonstrated the advantages of the framework on the VQA-CP V2 dataset through a number of experiments and analyses.

Zhang et al. propose a new VQA Kan that introduces richer visual information and compensates for common sense from an external knowledge base. For different types of problems, Zhang et al. make the model adaptively balance the importance of visual information and external knowledge. The self-adaptive scoring attention module can automatically select the appropriate information source according to the type of problem.

\subsection{Weakening language priors}
\label{ssec:observation}

Ramakrishnan et al. propose a new antagonistic regularization scheme to reduce the memory bias of data sets in VQA, which is based on the differences in the reliability of the model between the problem-only branches and the processed images. Experiments on VQA-CP datasets show that this technique allows existing VQA models to significantly improve performance in changing priorities. This approach can be implemented as a simple plug-in module on top of the existing VQA model, with end-to-end training from scratch.

Grand et al. examined several advantages and limitations of adversarial regularization, a recently introduced technique for reducing language bias in the VQA model. Although the authors find that AdvReg improves the performance of the outfield example in VQA-CP, one concern is that the pendulum swings too far: there are quantitative and qualitative indications that the model is over-normalized. The performance of the ADVREG model is influenced by the VQA-CP and the examples in the domain of the original VQA dataset. While AdvReg can improve performance for binary problems, it can degrade performance for other problem types. The author observes that ADVREG model makes use of significant image features and ignores important linguistic clues in the problem.

Cadene et al. proposed RUBi to reduce the unimodal bias learned by visual QA models. Rubi is a simple learning strategy designed to be model agnostic. It is based on a problem-only branch that captures unwanted statistics from the problem pattern. This branch influences the underlying VQA model to prevent learning single peak bias from problems. The authors show that the accuracy of VQA-CP V2 is significantly improved by + 5.94 percentage points compared with the latest results of VQA-CP V2, a data set used to explain problem deviations. Rubi is valid for different types of common VQA models.

Clark et al. propose a method that uses human knowledge about which methods can not be well generalized to improve the robustness of the model to domain transitions. This approach uses a pre-trained naive model to train the robust model in a set and a separate robust model for testing. A large number of experiments show that the method works well on two antagonistic datasets and two ever-changing prior datasets, including 12-point gain on VQA-CP.

KERVADEC and others have shown that properly structuring the semantic space of the output class can overcome some of the shortcomings of the widely used classification strategy in VQA, and their aim is to continue to pave the way for the long-term goals of the community, implementing a direct structured prediction of text output will bring VQA closer to the more traditional models of NLP.

Classical regularifiers applied to multi-modal datasets cause the model to ignore one or more modes. This is suboptimal because the authors expect all patterns to contribute to classification. To address this concern, Gat et al. studied regularization through functional entropy. It encourages a more uniform use of existing models.

To address the language bias issue, Niu and others argue that biased data sets should not be subject to time consuming depolarization. If the prediction made by the training model is biased, the reasons for the bias can be explored. When the model has only one input, I will only consider the bias data to get the predicted results. If the total effect (TE) is subtracted from the natural direct effect (NDE) , the language bias in the model prediction can be eliminated. In this framework, the authors use counterfactual reasoning to eliminate the effects of bias.

Han et al analyzed several methods of robust VQA through experiments, and proposed a new framework to reduce the language deviation in VQA. The author proves that language bias in VQA can be decomposed into distribution bias and shortcut bias, and proposes a greedy gradient integration strategy to eliminate these two biases. The experimental results show the rationality of the deviation decomposition and the effectiveness of GGE. The authors believe that the ideas behind GGE are valuable and have the potential to become a general method for solving data set bias problems. The authors extend GGE to solve other bias problems, provide more rigorous analysis to ensure convergence of the model, and learn to automatically detect different types of bias features without prior knowledge.

\subsection{Data argumentation and Training strategies}
\label{ssec:observation}

Teney et al proposed a new VQA method, which trains the interface between the model and the external data source and uses it to support its reply process. The model introduces new features, especially the use of non-VQA data to support the reply process. This provides many opportunities for future research to access “Black box”data sources such as web search and dynamic databases. This opens the door to systems that can reason about vision and language beyond the limited areas covered by any given training set.

Chen et al proposed a model unknown counterfactual synthesis (CSS) training scheme to improve the visual interpretation ability and problem sensitivity of the model. CSS generates counterfactual training samples by masking key objects or words. CSS continues to improve the performance of different VQA models. Chen et al. verified the effectiveness of CSS by extensive contrast and ablation experiments.

In order to make full use of the supervisory information of synthetic counterfactual samples in robust VQA, Liang and others introduced a self-supervised contrast learning mechanism to learn the relationship between fact samples and counterfactual samples. Experimental results show that this method improves the reasoning ability and robustness of VQA model.

At present, there are few researches on the internal causes of the language apriori problem in VQA. Guo et al. propose to fill this gap by explaining it from the perspective of class imbalance. Guo et al put forward two hypotheses and verified their feasibility. Guo et al have developed a simple and effective method of loss rescaling, which assigns different weights to each answer according to the given type of question. A large number of experiments on three public data sets verify the effectiveness of the proposed method. Guo and others hope to develop other approaches typically used to overcome class balancing problems, namely data rebalancing or transfer learning, to ease language prioritization.

Gokhale et al. proposed a method to train VQA model by using input mutation, which aims at realizing distributed generalization. Parallel work in the field of image classification shows that well-designed input perturbations or operations are beneficial to generalization and lead to performance improvements. Zhu et al proposed a new self-supervised learning framework to overcome language priors in VQA. Zhu and others argue that this work could be a meaningful step towards a realistic VQA and language bias, and that such self-monitoring could be extended to other tasks affected by inherent data bias. Experimental results show that the method achieves a balance between answering questions and overcoming linguistic prior knowledge.

Teney et al. ensure that data scaling uses exactly the same problem in each small batch, so improvements are strictly caused by architectural differences in approach. Under the condition of low training data, the improvement of this method is maximum. They show advantages under a variety of conditions, including distributed test data, low data training.

Guo et al proposed to solve the language apriori problem in VQA from the perspective of answer feature space learning, which has not been explored in previous studies. To achieve this goal, they designed an adaptive marginal cosine loss to distinguish the answer by correctly describing the answer feature space. For a given question, the frequent and sparse answers to the corresponding question types span wider and narrower angular spaces, respectively.

Yang et al. proposed a learning strategy called CCB to deal with language bias in visual question-and-answer (VQA) tasks. Based on the content and context branches, CCB guides the model by combining decisive content information and necessary context information. They constructed an additional loss function by separating the effects of the language bias on the model and jointly optimizing the two branches and the final prediction.

Jiang et al. proposed a graph-based generation modeling scheme X-GGM to improve the OOD generalization ability of baseline VQA model while maintaining ID performance. The X-GGM scenario randomly executes R-GGM or N-GGM to generate a new relational matrix or a new node representation. The gradient derivative and disturbance distribution are used to solve the problem of unstable distribution of adversarial data.

Yuan et al. proposed a counterfactual thinking module to solve the problem of language bias, and helped the model improve the counterfactual thinking ability through multi-level construction of comparative learning loss. It alleviates the problem of language bias from the perspective of causal inference and data enhancement.

\iffalse
\begin{figure}[!t]
\def\xheight{0.823}
\centering
\includegraphics[width=\xheight\linewidth]{figs/relative_range.pdf}
\caption{Relative range $\mathrm{R}^k_i$ comparisons of the selected six features calculated on the three UGC-VQA databases: KoNViD-1k, LIVE-VQC, and YouTube-UGC.}
\label{fig:relative_range}
\end{figure}

\begin{figure}[!t]
\def\xheight{0.823}
\centering
\includegraphics[width=\xheight\linewidth]{figs/coverage_uniformity.pdf}
\caption{Comparison of coverage uniformity $\mathrm{U}^k_i$ of the selected six features computed on the three UGC-VQA databases: KoNViD-1k, LIVE-VQC, and YouTube-UGC.}
\label{fig:uniformity}
\end{figure}
\fi

\section{Datasets and evaluation}
\label{sec:nr_vqa}

\indent 
In the field of VQA, researchers have proposedd many datasets to evaluate the performance of VQA. These data sets generally contain a triple consisting of an image, a question and the correct answer. Sometimes, these datasets provide additional annotation information. Such as image title, image area and multiple candidate answers.
\subsection{Datasets}
At present, researchers usually use VQA CP V1 and V2 data sets to evaluate the performance of the proposedd model. And perform auxiliary verification on the data set VQA V1 or V2. Most of the existing research results are tested on VQA CP V2, and a few will also be tested on vqacp v1. The training set of vqacpv2 dataset contains 121k pictures, 438k questions and 4.4m answers, and its test set contains 98K pictures, 220K questions and 2.2m answers. The dataset is divided by VQA V2 dataset. In order to measure the problem of language bias, the question and answer distribution of the training set and the test set of the data set are quite different, that is, for the same type of questions, the training set and the test set have inconsistent answer distribution. Therefore, the data set is very suitable to measure the generalization ability and robustness of the model.

\noindent\textbf{\emph{VQA v1:}}
VQA v1 is the first version of the VQA dataset. The training set contains 248349 image problem pairs, the verification set contains 121512 image problem pairs, and the test set contains 244302 image problem pairs. In the test of the model, the annotation of the test set is not available except for the remote server used for evaluation. All images of VQA V1 are from the mscoco dataset. For each sample pair in the data set, there is a question, an image and ten ground-truth answers.

\noindent\textbf{\emph{VQA v2:}}
Vqa-2.0 is the second version of VQA data set. The training set contains 443757 image problem pairs, the verification set contains 214354 image problem pairs, and the test set contains 447793 image problem pairs. Its dataset is twice as large as its first version. The questions and answers in the data set are also artificially annotated. In addition, for each question in the dataset, there are two similar images, but the answers are different, which makes the sample distribution of the dataset more balanced. Vqa-2.0 datasets are larger and more balanced than vqa-1.0.

\noindent\textbf{\emph{VQA-CP v1:}}
VQA-CP v1    dataset is obtained after sample re division from VQA V1 data set. The training set of vqacp V1 data set contains 118K pictures, 245k questions and 2.5m answers, and its test set contains 87k pictures, 125k questions and 1.3m answers.

\noindent\textbf{\emph{VQA-CP v2:}}
VQA-CP v2 dataset is obtained after sample redivision from VQA v2 dataset. The training set of vqacp V2 data set contains 121k pictures, 438k questions and 4.4m answers, and its test set contains 98K pictures, 220K questions and 2.2m answers. This data set is the most used data set for testing methods by researchers.

%其中，我们展示了基于UPDN模型的方法和其对应的结果。
\begin{table*}[ht]
\centering
\caption{\textbf{Comparison on VQA-CP v2 test set and VQA v2 val set}. We show the method based on the UpDn model and its corresponding results. ``Base.'' indicates the VQA base model.}
\label{tab:all}
\vspace{-2mm}
\scalebox{0.93}{
\begin{tabular}{l c c c cccc c cccc}
\hline
\toprule
Test set & & & & \multicolumn{4}{c}{VQA-CP v2 test} & & \multicolumn{4}{c}{VQA v2 val}\\  
\cmidrule{5-8} \cmidrule{10-13}
Methods & ~ & Base. & ~ & {All} & {Y/N} &  {Num.} & {Other} & ~ & {All} & {Y/N} & {Num.} & {Other}\\
\midrule
GVQA~\cite{agrawal2018don}   &  & -- & &  31.30 & 57.99 & 13.68  & 22.14 & & 48.24 & 72.03   & 31.17  & 34.65 \\
SAN~\cite{yang2016stacked}  &  & -- & &   24.96 & 38.35 & 11.14  & 21.74 & & 52.41 & 70.06 & 39.28 & 47.84 \\
UpDn~\cite{anderson2018bottom}  &  & -- & & 39.74 & 42.27 & 11.93 & 46.05 & & 63.48 & 81.18 & 42.14 & 55.66 \\
S-MRL~\cite{cadene2019rubi} & & -- & &  38.46 & 42.85 & 12.81 & 43.20 & & 63.10 & -- & -- & -- \\
NSM~\cite{hudson2019learning}   &  & -- & &  45.80 & -- & --  & -- & & -- & --   & --  & -- \\
MUREL~\cite{cadene2019murel}   &  & -- & &  39.54 & 42.85 & 13.17  & 45.04 & & 65.14 & --   & --  & -- \\
CAM~\cite{peng2020answer}   &  & -- & &  39.75 & 43.29 &  12.31  & 45.41 & & -- & --   & --  & -- \\
LXMERT~\cite{tan2019lxmert}& & -- & &  41.28 & 42.01 & 14.16 & 48.34 & & 65.31 & 83.30 & 46.15 &56.91 \\
% a pitcure 
% 46.23 42.84 18.91 55.51 74.16 89.31 56.85 65.14 acl check it again

\hline
\multicolumn{13}{l}{\it methods based on strengthening visual information:} \\
\hline
AttAlign~\cite{selvaraju2019taking} & & UpDn & &  39.37 & 43.02 &11.89 & 45.00 & & 63.24 & 80.99 & 42.55 & 55.22 \\
HINT~\cite{selvaraju2019taking} & & UpDn & &  46.73 & 67.27 & 10.61 & 45.88 & & 63.38 & 81.18 & 42.99 &  55.56 \\
SCR~\cite{wu2019self} & & UpDn & & 49.45 & 72.36 & 10.93 & 48.02 & & 62.2 & 78.8 & 41.6 & 54.5 \\
ReGAT~\cite{li2019relation} & & UpDn & &  40.42 & -- &-- & -- & & 67.18 & -- & -- & -- \\
ESR~\cite{shrestha2020negative} & & UpDn & &  48.9 & 69.8 & 11.3 & 47.8 & & 62.6 & -- & -- & -- \\
VGQE~\cite{kv2020reducing} & & UpDn & &  48.75 & -- &-- & -- & & 64.04 & -- & -- & -- \\
Picture~\cite{hirota2021picture} & & UpDn & &  43.64 & 45.13  & 20.06 & 49.33 & & 69.74 & 87.91 & 56.47 & 59.43 \\
SAR~\cite{si-etal-2021-check} & & UpDn & & 61.71 & --& -- & -- & & -- & -- & -- & -- \\
KAN~\cite{zhang2020rich} & & UpDn & & 42.60 & 42.12 & 15.52 & 50.28 & & -- & -- & -- & -- \\

\hline
\multicolumn{13}{l}{\it methods based on weakening language priors:} \\
\hline
AdvReg~\cite{ramakrishnan2018overcoming} & & UpDn & &  41.17 & 65.49 & 15.48 & 35.48 & & 62.75 & 79.84 & 42.35 & 55.16 \\
GRL~\cite{grand2019adversarial} & & UpDn & & 42.33 & 59.74 & 14.78 & 40.76 & & --  &  --  & --  & -- \\
RUBi~\cite{cadene2019rubi} & & UpDn & & 44.23 & 67.05 & 17.48 & 39.61 & & --  &  --  & --  & -- \\
LM~\cite{clark2019don} & & UpDn & & 48.78 & 72.78 & 14.61 & 45.58 & & 63.26 & 81.16 & 42.22 & 55.22 \\
LM+H~\cite{clark2019don} & & UpDn & & 52.01 & 72.58 & 31.12 & 46.97 & &56.35 & 65.06 & 37.63 & 54.69 \\
Semantic~\cite{kervadec2020estimating} & & UpDn & &  47.5 & -- & -- & -- & & -- & -- & -- & -- \\
RMFE~\cite{gat2020removing} & & UpDn & & 54.55 & 74.03 & 49.16 & 45.82 & & -- & -- & -- & -- \\

CF-VQA~\cite{niu2021counterfactual} & & UpDn & & 53.55$^{\pm 0.10}$ &  91.15$^{\pm 0.06}$ & 13.03$^{\pm 0.21}$ & 44.97$^{\pm 0.20}$ & & 63.54$^{\pm 0.09}$ & 82.51$^{\pm 0.12}$ & 43.96$^{\pm 0.17}$ & 54.30$^{\pm 0.09}$ \\
GGE-DQ~\cite{han2021greedyiccv} & & UpDn & & 57.32 & 87.04 & 27.75 & 49.59 & &59.11 & 73.27 & 39.99 & 54.39 \\
LPF~\cite{liang2021lpf} & & UpDn & & 55.34 & 88.61 & 23.78 & 46.57 & &55.01 &64.87 & 37.45 & 52.08 \\

%\midrule
\hline
\multicolumn{13}{l}{\it methods based on data argumentation and training strategies:} \\
\hline
ActSeek~\cite{8953570} &  & UpDn &  &  46.00 & 58.24 & 29.49 & 44.33 & & -- & -- & -- & -- \\
A1C-WS~\cite{zhou2019plenty} &  & UpDn &  &  39.6 & 42.7 & 12.9 & 45.3  & & -- & -- & -- & -- \\
CSS~\cite{chen2020counterfactual} & & UpDn &  & 58.95 & 84.37 & 49.42 & 48.21
 & & 59.91 & 73.25 & 39.77 & 55.11 \\
CL-VQA~\cite{liang2020learning} & & UpDn &  & 59.18 & 86.99 & 49.89 & 47.16 & & 57.29 & 67.27 & 38.40 & 54.71 \\
GradSup~\cite{teney2020learning} & & UpDn &  & 46.8 & 64.5 & 15.3 & 45.9
 & & -- & -- & -- & -- \\
Loss-Rescaling~\cite{guo2020loss} & & UpDn &  & 53.26 & 72.82 & 48.00 & 44.46 & & 56.81 & 68.21 & 36.37 & 52.29 \\
Mutant~\cite{gokhale2020mutant} & & UpDn &  &  61.72
 & 88.90 & 49.68 & 50.78
 & & 62.56 & 82.07 & 42.52 & 53.28 \\
RandImg~\cite{NEURIPS2020_045117b0} &  & UpDn &  &  55.37 & 83.89 & 41.60 & 44.20 & & 57.24 & 76.53 & 33.87 & 48.57 \\
SSL~\cite{ijcai2020-151} & & UpDn &  & 57.59 & 86.53 & 29.87 & 50.03
 & & 63.73 & -- & -- & -- \\
Unshuffling~\cite{teney2021unshuffling} &  & UpDn &  &  42.39 & 47.72 & 14.43 & 47.24  & & 61.08 & 78.32 & 42.16 & 52.81 \\
LP-Focal~\cite{lao2021language} &  & UpDn &  &  58.45 & 88.34 & 34.67  & 49.32 & & 62.45 & -- & -- & -- \\
ADA-VQA~\cite{ijcai2021-98} & & UpDn &  & 54.67 & 72.47 & 	53.81 & 45.58
 & & -- & -- & -- & -- \\
CCB-VQA~\cite{yang2021learning} & & UpDn &  & 59.12& 89.12& 51.04 &45.62& & 59.17 &77.28 &33.71 &52.14 \\
SBS~\cite{ouyang2021suppressing} & & UpDn &  & 59.57 & 87.44 & 52.96 & 46.79 & & 61.97 & 78.80 & 42.17 & 54.41 \\
WeaQA~\cite{banerjee2021weaqa} & & UpDn &  &  41.2
 & 68.5 & 29.8 & 30.0
 & & -- & -- & -- & -- \\
X-GGM~\cite{jiang2021x} & & UpDn &  &  45.71
 & 43.48 & 27.65 & 52.34
 & & -- & -- & -- & -- \\
CFT-VQA~\cite{yuan} & & UpDn &  &  59.37
 & 87.95 & 52.42 & 46.30
 & & 59.82 & 74.91 & 38.64 & 53.97 \\
\bottomrule
\end{tabular}
}
\end{table*}

\subsection{Evaluation Measures}

How to evaluate the correctness and rationality of sentences generated by computer is a basic and difficult task. When evaluating the correctness of sentences, we need to consider the correctness of syntax and whether the semantic information expressed by sentences is correct. In order to simplify this problem, most data sets of VQA are evaluated by limiting the generated answer sentences to a single word or phrase, generally 1 to 3 words in length. Through simplification, the semantic expression of sentences is more specific, and the difficulty of sentence correctness evaluation is greatly reduced.

Malinowski et al. creatively proposedd two indicators for VQA tasks. The first method uses string to match the difference between the generated sentence and ground truth. It can be considered correct only when the two sentences are completely consistent. The second method uses Wu Palmer similarity to evaluate the generated sentences. This method evaluates the similarity between the common subsequences of the two sentences in the syntax classification tree. When the similarity between two lists exceeds a fixed threshold, the generated sentence (answer) is considered to be correct.

VQA-real dataset collected 10 different ground truth answers for each question in order to solve the problem of fuzzy evaluation criteria in VQA evaluation. When evaluating this data set, the generated sentences (answers) need to be compared with the answers of 10 human answers. The formula is as follows:
\begin{equation}
accuracy =\min \left(\frac{\# \text { humans provided that answer }}{3}, 1\right)
\end{equation}

\subsection{Results of existing methods}

At present, most of the existing methods to solve the problem of language bias in VQA are evaluated on VQA-CP v2 and verified on VQA v2. This paper investigates the papers published in top conferences and relevant journals in recent four years, and counts the existing experimental results in table 1. These results are derived from their original papers.

\section{Discussion and future directions}
\label{sec:fs}

VQA is a relatively new task compared with other visual tasks. It has produced great development in just a few years, but it is far from mature compared with other tasks, such as image recognition, face recognition and other tasks that can be almost completely applied. Compared with VQA's initial vision and completely free visual QA in a completely open world, the current VQA task adds many constraints and restrictions to simplify this task. At the same time, due to the existence of language bias, there are great difficulties and limitations in the application of visual question answering. Therefore, this paper summarizes the progress of deep learning and existing methods, and gives possible future research directions.

\noindent\textbf{\emph{External knowledge:}} VQA task is a complete artificial intelligence task, which contains image information and language information to realize intelligent question and answer. However, the current VQA task is generally to find the correct answer from the preset answer. One of the sources of language bias is due to the uneven distribution of data and question answers. Therefore, for a reasonable VQA system, it should search for the correct answers more widely from the external database, rather than using the predetermined question types and answer types. This has great limitations on the application of VQA. Using more complete external data sets or combining knowledge maps to answer questions may be one of the effective means to reduce language bias, which can help the model obtain a more evenly distributed question answer database. At present, there are some data sets related to external databases\cite{cao2021knowledge, marino2019ok, ziaeefard2020towards} in other VQA tasks, but the existing methods generally improve performance through unlimited cumulative data set size. How to find more efficient external knowledge assistance methods is one of the research directions to help solve language bias.

\noindent\textbf{\emph{Long-tail learning:}} Because the distribution of question answers in the data set is uneven, there is a long tail distribution problem. However, the labels of questions and answers are uniformly divided into counting, whether or not. The answers are simply classified and contain less bias information. The category is a banana color problem. Due to the lack of more fine-grained banana color labels, it is difficult to simply transform the language bias problem into a long tail\cite{menon2020long, wang2020one,ouyang2016factors,zhou2018deep} distribution problem. This is also reasonable in real life. Due to the limitations of manpower and resources, it is difficult to classify data sets very fine-grained, which inevitably leads to the imbalance of more fine-grained labels within the classification, and also the problem of language bias. Therefore, how to resolve the language bias problem in VQA into a long tail distribution problem, simplify the problem and realize the in-depth mining of language and visual information to extract fine-grained labels may be the most effective research direction.

\noindent\textbf{\emph{Causality:}} Some of the existing methods to slow down language priors and data enhancement can be classified as causal inference methods\cite{hernan2010causal, holland1986statistics, pearl2009causal,pearl2010causal,morgan2015counterfactuals,pearl2016causal}. The problem of language bias in VQA can be analyzed from the perspective of causality, including the construction of causality diagram~\cite{niu2021counterfactual}, counterfactual data enhancement~\cite{chen2020counterfactual} and counterfactual thinking~\cite{yuan}. In the case of balanced distribution of data sets, the causality method can help the model better analyze and use the core information of images and problems, and correlate the two. Counterfactual thinking is a method of modeling human thinking process. Through targeted data enhancement and constructing counterfactual examples to deduce the model from multiple levels, we can analyze the source of language bias from the perspective of causality and slow it down to a certain extent. In most cases, there is no bias due to different data. Therefore, how to let the machine learn the profound causal relationship between data and learn the way of human thinking is an effective way to solve the language bias.

\section{Conclusion}
\label{sec:conclud}
As the first survey in this field, this paper makes a comprehensive review of language bias in Visual Question Answering (VQA). This paper combs the existing methods in detail, which is divided into three parts: enhancing visual information, reducing language information, data enhancement and training processing. We introduce relevant methods and re-examine the advantages and disadvantages of existing methods. At the same time, the mainstream data sets and the experimental results of various methods on the data sets are also introduced. In addition, we have not pointed out some possible hopes for the future development and research in this field, and revealed that language bias is a common phenomenon in practical application and deep learning.

% or


% \appendices

% % you can choose not to have a title for an appendix
% % if you want by leaving the argument blank
% \section{}

% % use section* for acknowledgment
% \section*{Acknowledgment}
% \label{sec:ack}


\ifCLASSOPTIONcaptionsoff
  \newpage
\fi



%\bibliographystyle{IEEEtran}

\bibliographystyle{IEEEtran}
\bibliography{ref}

\end{document}